\documentclass[letterpaper]{article}
\pdfoutput=1
\usepackage{uai2018}
\usepackage[margin=1in]{geometry}

\usepackage{times}

\usepackage{microtype}
\usepackage{graphicx}
\usepackage{subfigure}
\usepackage{booktabs} 
\usepackage{dsfont} 
\usepackage{multirow}
\usepackage{centernot}
\usepackage{enumitem}
\setlist{itemsep=0.01cm,topsep=0.02cm,leftmargin=*}
\usepackage{afterpage}

\usepackage{amsmath,amssymb,amscd,amsfonts,amsthm,mathtools}

\usepackage[round,authoryear]{natbib}
\usepackage{hyperref}
\usepackage{cleveref}

\usepackage[T1]{fontenc}

\crefname{section}{Section}{Sections}
\crefname{table}{Table}{Tables}
\crefname{figure}{Figure}{Figures}
\crefname{appsec}{Appendix}{Appendices}

\theoremstyle{plain}

\theoremstyle{definition}

\newcommand{\calV}{\mathcal{V}}
\newcommand{\calE}{\mathcal{E}}
\newcommand{\ee}{Z} 
\newcommand{\bfee}{\mathbf{\ee}}
\newcommand{\bfE}{\mathbf{E}}
\newcommand{\bfG}{\mathbf{G}}
\newcommand{\PYP}{\mathcal{PYP}}
\newcommand{\geom}{\beta}
\newcommand{\BNTL}{\text{\rm BNTL}}
\newcommand{\bfT}{\mathbf{T}}
\newcommand{\calO}{\mathcal{O}}
\newcommand{\bbR}{\mathbb{R}}
\newcommand{\bfP}{\mathbf{P}}
\newcommand{\bfPsi}{\boldsymbol{\Psi}}
\newcommand{\bfn}{\mathbf{d}}

\def\iid{i.i.d.\ }

\newcommand{\argdot}{{\,\vcenter{\hbox{\scalebox{0.5}{$\bullet$}}}\,}}

\def\bbE{\mathbb{E}}
\def\indicator{\mathds{1}}

\def\Poisson{\text{\rm Pois}}

\def\BetaDist{\text{\rm Beta}}

\def\Geom{\text{\rm Geom}}

\newcommand\Categorical{\text{\rm Categorical}}

\def\randperm{\sigma}

\def\randPart{\Pi}

\def\aTime{T}

\def\aDist{\Lambda}

\def\bbN{\mathbb{N}}
\def\bbP{\mathbb{P}}

\newcommand{\widesim}[1][1.5]{
  \mathrel{\scalebox{#1}[1]{$\sim$}}
}
\newcommand{\limscale}[2]{\overset{\scriptscriptstyle{#1 \uparrow #2}}{\widesim[1.25]}}

\DeclareMathOperator*{\argmax}{arg\,max}

\title{Sampling and Inference for Beta Neutral-to-the-Left Models of Sparse Networks}

\author{ {\bf Benjamin Bloem-Reddy} \\
Department of Statistics \\
University of Oxford\\
\And
{\bf Adam Foster}  \\
Department of Statistics \\
University of Oxford\\
\And
{\bf Emile Mathieu}   \\
Department of Statistics \\
University of Oxford\\
\And
{\bf Yee Whye Teh} \\
Department of Statistics \\
University of Oxford
}

\begin{document}

\maketitle

\begin{abstract}
Empirical evidence suggests that heavy-tailed degree distributions occurring in many real networks are well-approximated by power laws with exponents $\eta$ that may take values either less than and greater than two. Models based on various forms of exchangeability are able to capture power laws with $\eta < 2$, and admit tractable inference algorithms; we draw on previous results to show that $\eta > 2$ cannot be generated by the forms of exchangeability used in existing random graph models.  Preferential attachment models generate power law exponents greater than two, but have been of limited use as statistical models due to the inherent difficulty of performing inference in non-exchangeable models. 
Motivated by this gap, we design and implement inference algorithms for a recently proposed class of models that generates $\eta$ of all possible values. We show that although they are not exchangeable, these models have probabilistic structure amenable to inference. Our methods make a large class of previously intractable models useful for statistical inference. 
\end{abstract}

\section{INTRODUCTION}
\label{sec:intro}

Sparsity and heavy-tailed degree distributions are believed to occur in many real networks \citep{Newman:2005,Clauset:Shalizi:Newman:2009}. Sparsity has been well-studied and is an intuitive concept: The typical social network user interacts with only a vanishing fraction of all users as the network grows. Heavy-tailed degree distributions and the mechanisms that generate them are not as well understood. However, empirical evidence indicates that heavy-tailed distributions are expressed in a wide range of settings, including network degrees \citep{Clauset:Shalizi:Newman:2009}. 
Power law degree distributions, in which the proportion of vertices with degree $d$ is $\propto d^{-\eta}$, are often used as models for real degree distributions, and serve as a useful analytic tool for characterizing the asymptotic properties of random network models. 

Many statistical network models make the assumption of exchangeability over vertices, appealing to the Aldous--Hoover theorem \citep{Hoover:1979,Aldous:1981} for theoretical justification. 
Noting that networks sampled from these models cannot be sparse, \citet{Orbanz:Roy:2015} posed a question paraphrased as, ``Can a probabilistic model for random graphs produce sparse networks and have some useful notion of probabilistic symmetry?'' A generation of models answered in the affirmative by incorporating other notions of exchangeability: In an exchangeable point process representation of a network \citep{Caron:Fox:2017,Veitch:Roy:2015,Borgs:etal:2016}, or as an exchangeable sequence of edges \citep{Crane:Dempsey:2017,Cai:etal:2016,Williamson:2016}. Under certain parameterizations, these models generate sparse networks. They are able to generate asymptotic power law degree distributions, providing a better fit to real network data than their vertex exchangeable counterparts. However, the power law exponent of the degree distribution in both model classes is constrained to the interval $\eta\in(1,2)$. That interval is not an artifact of particular model specifications. Rather, it is a basic property resulting from the fact that the average vertex degree is asymptotically unbounded; vertex degrees grow, on average, linearly in the number of edges. For some data, this property may be undesirable; such properties ideally would be inferred from a model able to capture a larger range of power law behavior. 

In a largely disjoint literature, so-called preferential attachment (PA) models have been studied primarily for their ability to generate power law degree distributions from a simple size-biased reinforcement mechanism, and for their analytical tractability \citep[e.g.,][]{Barabasi:Albert:1999,Berger:etal:2014,Pekoz:Rollin:Ross:2017}. PA models have power law exponents $\eta>2$. As we explain in \cref{sec:powerlaws}, the exponent range is tied to PA models' non-exchangeability---a property that has made them, until now, of limited use as statistical models. In particular, if the history of the network is unobserved, the order of the edges must be inferred or marginalized; even for networks of modest size, such inference over permutations is generally intractable.

Recently, \citet{Bloem-Reddy:Orbanz:2017aa} introduced a class of models that can generate random graphs with power law degree distributions of any exponent $\eta\in(1,\infty)$. For reasons discussed below, we propose naming them Beta Neutral-to-the-Left (BNTL) models. BNTL models generalize many known models that have a size-biased reinforcement mechanism, including a sub-class of edge exchangeable models based on the Pitman--Yor process, and variations of the PA model. The cost of the additional flexibility is exchangeability; BNTL models depend on the times at which new vertices arrive and are not exchangeable in any known sense.  
However, as we show in \cref{sec:betaNTL}, BNTL models have probabilistic structure---namely, left-neutrality---that may be exploited for efficient computation, making a large class of previously intractable models useful for statistical inference. 

\Citet{Bloem-Reddy:Orbanz:2017aa} established the asymptotic properties of BNTL models; statistical modeling and inference were left unstudied. Our contributions are:
\begin{itemize}
  \item We identify left-neutrality as the key property that yields tractable inference schemes.
  \item We categorize and give solutions to the BNTL inference problem based on what data are available: We design schemes for maximum likelihood estimation when vertex arrival times are observed, and for Bayesian inference when an unlabeled network is observed.
  \item We implement these schemes on real networks of various sizes, from modest ($\sim 10^2$ vertices) to massive ($\sim 10^6$ vertices).
\end{itemize}

\section{POWER LAWS IN RANDOM GRAPH MODELS}
\label{sec:powerlaws}

This section provides some context, and collects and interprets various results from random graph models with asymptotic degree distributions exhibiting power law tails. Although none of the results here are new, to our knowledge they have not been coherently synthesized in the literature. Technical details are omitted; they may be found in the references given throughout the section. We focus our attention on edge exchangeable and PA models because they are most similar to the BNTL framework. 

A graph $G$ is a set of vertices, $\calV(G)$, and of edges,\footnote{We treat all graphs as undirected; extension to directed graphs is straightforward.} $\calE(G)$, between them. A multigraph allows for multiple edges between vertices; we consider each edge to be distinct, rather than as one integer-valued edge. We consider only multigraphs and henceforth refer to them as graphs. A sequence of growing graphs $G_1,G_2,\dotsc$ is a stochastic process $\bfG$, indexed by the number of edges, $n$. Hence, $G_{n}$ may be interpreted as $G_{n-1}$ with an additional edge, either between two vertices in $\calV(G_{n-1})$, between to new vertices, or between one old and one new vertex. We assume that the edges are labeled according to the order in which they appear, though this assumption is not necessary for edge exchangeable models (discussed below). As such, $G_n$ may be viewed simply as a sequence of edges $\bfE_n:=(E_1,\dotsc,E_n)$ or, even more simply, as a sequence of \emph{ends of edges}, denoted ${\mathbf{\ee}_{2n} = (\ee_1,\dots,\ee_{2n})}$. We denote by $G(\bfee_{2n})$ the labeled graph with $n$ edges constructed from $\bfee_{2n}$. (For convenience, we will use the subscript $n$ for all sequences when there is no risk of confusion.)

For a graph $G(\bfee_n)$, $K_{n} :=|\calV(G(\bfee_n))|$ is the number of vertices (i.e., the number of unique values in $\bfee_n$); the degree of vertex $j$, $d_{j,n} := \sum_{i=1}^n \indicator\{\ee_i=j\}$, is equal to the number of ends of edges connected to it. Let $m_{n}(d)$ denote the number of vertices with degree $d$. The asymptotic degree distribution of $G_1,G_2,\dotsc$ is said to have power law tail with exponent $\eta > 1$ if
\begin{align*} 
  p_{n}(d) = \frac{m_{n}(d)}{K_n} \xrightarrow[n\to\infty]{p} p_d \limscale{d}{\infty} L(d)d^{-\eta} \quad \text{for large $d$} \;,
\end{align*}
such that $\sum_{d\geq 1} p_d =1$, for some slowly varying function $L(d)$: ${\lim_{x\to\infty} L(rx)/L(x) = 1}$ for all ${r>0}$ \citep{Bingham:1989}. 
For power law tails, we state the following fact (see \Cref{sec:app:power:laws}). 


\noindent\emph{Fact.} As $n\to\infty$, if the expected average degree is unbounded, then $\eta \in (1,2)$; if it is bounded, ${\eta\in (2,\infty)}$.

\noindent\textbf{Edge exchangeable models \citep{Crane:Dempsey:2017,Cai:etal:2016}.} Let $G_n$ be specified by its sequence of edges $\bfE_n$ (not necessarily ends of edges), which is assumed to be exchangeable: Its distribution is invariant under all permutations of the order of the edges for all $n$, i.e., the labels carry no information about their distribution. As a consequence of the law of large numbers for exchangeable sequences, the counts of all non-zero multi-edges grow linearly in $n$ and thus so do the vertex degrees. That is, $d_{j,n} = \Theta(n)$. Furthermore, $K_n = o(n)$. The average degree is unbounded, implying that if the degree distribution tail follows a power law, then $\eta\in(1,2)$. 

As an example, consider sampling $\bfee$ from the the Pitman--Yor process ($\PYP$) \citep{Ishwaran:James:2001} with parameters $\tau \in (0,1)$, $\theta > -\tau$,
\begin{align} \label{eq:pyp}
  \bbP[\ee_{n+1} \in \argdot | \bfee_{n}] & = \frac{\theta + K_n \tau}{n + \theta} \delta_{K_n + 1}(\argdot) \\
  & \quad + \frac{n - K_n\tau}{\theta + n}\sum_{j=1}^{K_n} \frac{d_{j,n} - \tau}{n - K_n\tau}\delta_{j}(\argdot) \nonumber \;.
\end{align}
It can be shown that the asymptotic degree distribution has power law tail \citep{Pitman:2006},
\begin{align*}
  n^{-\tau} m_d(n) \xrightarrow[n\to\infty]{\text{\small a.s.}} p_d \limscale{d}{\infty} d^{-(1 + \tau)} \;,
\end{align*}
which implies that $\eta_{\tau}=1+\tau\in (1,2)$.

The predictive rule \eqref{eq:pyp} demonstrates why the expected average degree is unbounded. The probability that $\ee_{n+1}$ corresponds to a new vertex is $\frac{\theta + K_n\tau}{n + \theta}$, which is arbitrarily close to zero as $n\to\infty$. For large $n$, the expected interarrival time between new vertices becomes arbitrarily large, and edges pile up on the existing vertices. Intuitively, vertex $j$ takes part in a constant fraction of all interactions as $n$ grows. This property is shared by all edge exchangeable models; an analogous property holds for exchangeable point process models (see \Cref{sec:app:graphex}).

\noindent\textbf{Preferential attachment models.} Although the $\PYP$ has the same size-biased reinforcement mechanism common to all PA models, typically it is not considered to be part of the same class as the PA models in the probability literature, of which \citet{Barabasi:Albert:1999} provide the prototypical example. However, the difference between them amounts to how frequently new vertices appear \citep{Bloem-Reddy:Orbanz:2017aa}. For our purposes, this is best illustrated with a simple PA model, the Yule--Simon (YS) model \citep{Simon:1955}. For $\geom\in(0,1)$, $\bfee$ is generated via the predictive rule
\begin{align}  \label{eq:ys}
  \bbP[\ee_{n+1} \in \argdot | \bfee_{n}] & = \geom \delta_{K_n + 1}(\argdot) + (1-\geom) \sum_{j=1}^{K_n} \frac{d_{j,n}}{n} \delta_j(\argdot) .
\end{align}
The YS model is known to generate power law degree distributions with $\eta_{\geom} = 1 + \frac{1}{1-\geom}\in(2,\infty)$ \citep{Simon:1955}. Different versions of PA exhibit a range of possible $\eta$'s, but it is generally the case that $\eta_{\text{\tiny PA}} > 2$, and this is tied to their lack of exchangeability. The average rate at which new vertices arrive is constant in $n$; hence, $K_n = \Omega(n)$, implying bounded expected average degree. The ``edge pileup'' phenomenon of exchangeable models does not occur: $d_{j,n}=o(n)$. In the YS model, $d_{j,n}=\Theta(n^{1-\geom})$.

\section{BETA NTL MODELS}
\label{sec:betaNTL}

BNTL models were introduced under the name $(\alpha,T)$-models by \citet{Bloem-Reddy:Orbanz:2017aa}, who studied their distributional and asymptotic properties. We briefly review the definition of BNTL models and describe the properties that make them amenable to inference. 

In the predictive distributions \eqref{eq:pyp}-\eqref{eq:ys}, the probability that $Z_{n+1}$ is a new vertex is independent of the degrees $d_{j,n}$,
which allows the sampling of $\bfee$ to be separated into two parts: A sequence $T_1 < T_2 < \dotsc$ of \emph{arrival times} of new vertices, and size-biased reinforcement at all steps not associated with an arrival time. 
As such, a BNTL model is parameterized by a scalar ``discount parameter'' $\alpha\in(-\infty,1)$ and a probability distribution $\aDist$ on strictly increasing integer-valued sequences, which specifies the law of the arrival times $T_1,T_2,\dotsc$. 
A sequence $\bfee$ is said to have law $\BNTL(\alpha,\aDist)$ if, for a random arrival time sequence $\bfT=(T_1,T_2,\dotsc)\sim\aDist$, $\bfee$ is sampled as 
\begin{align} \label{eq:bntl}
  \bbP & [\ee_{n+1} \in \argdot | \bfee_{n}, \bfT] = \indicator\{ n+1=T_{K_n + 1} \} \delta_{K_n + 1}(\argdot) \nonumber \\
  & \quad + \indicator\{ n+1<T_{K_n + 1} \} \sum_{j=1}^{K_n} \frac{d_{j,n}-\alpha}{n-K_n\alpha} \delta_j(\argdot)  \;.
\end{align}
In practice, it may be simpler to specify the distribution of \emph{interarrival} times ${\Delta_j = T_j - T_{j-1}}$, and use their partial sums to construct $\bfT$; we discuss this in more detail in \cref{sec:inference}. 
The similarity of \eqref{eq:bntl} to \eqref{eq:pyp}-\eqref{eq:ys} is not coincidental. The $\PYP$ and the YS model each correspond to particular parameterizations of the BNTL model: The YS model corresponds to \iid $\Delta_j\sim\Geom(\geom)$; the arrival time distribution induced by the $\PYP$ in \eqref{eq:pyp} also has known form (see \eqref{eq:arrivals:pyp}). 

For a given $\bfT$, the probability of any $G(\bfee_n)$ is
\begin{align} \label{eq:cppf}
  & \bbP[G(\bfee_n)|\bfT] = \bbP[G(\bfee_n)|\bfT_{K_n+1}, K_n]  \\ 
  & \; = \frac{\Gamma(d_{1,n}-\alpha)}{\Gamma(n - K_n\alpha)} \prod_{j=2}^{K_n} \frac{\Gamma(T_j - j\alpha) \Gamma(d_{j,n} - \alpha)}{\Gamma(T_j - 1 - (j-1)\alpha)\Gamma(1-\alpha)} \nonumber \;.
\end{align}
A crucial property that makes BNTL models amenable to inference is that conditioned on $\bfT$, the joint probability \eqref{eq:cppf} factorizes over the vertices; each term is expressed in terms of its arrival time, $T_j$, and its degree, $d_{j,n}$. Note that given $\bfT$, the degree sequence $\bfn_{K_n}:=(d_{1,n},\dotsc,d_{K_n,n})$ is a sufficient statistic for $\alpha$. Furthermore, the distribution of the arrival times (and therefore $K_n$) is independent of the degrees. The factorization becomes explicitly useful in the Gibbs sampling updates and in the maximum likelihood estimating equations in \cref{sec:inference}.

\noindent\textbf{Sampling representation.} Like their exchangeable counterpart the $\PYP$, BNTL models have a sampling representation in terms of products of independent beta random variables: 
\eqref{eq:bntl} is an urn sequence corresponding to the following \citep{Bloem-Reddy:Orbanz:2017aa}:
\begin{itemize}
  \item $\bfT \sim \aDist$.
  \item $\Psi_j | T_j \sim\BetaDist(1-\alpha,T_j-1-(j-1)\alpha)$ for $j\geq 1$.
  \item $P_{j,K_n} = \Psi_j\textstyle\prod_{\ell=j+1}^{K_n} (1-\Psi_{\ell}) $ 
  \item $\ee_n \sim 
    \begin{cases} 
      \delta_{K_n}(\argdot) & \text{ for } n=T_{K_n} \\
      \Categorical(P_{j,K_n}) & \text{ o.w. } 
    \end{cases}
    $
\end{itemize}
(By convention, $\BetaDist(a,0)$ is a point mass on 1, so $\Psi_1=1$.) 
The last two items specify that when there are $k$ vertices in the graph, $\ee_n$ is sampled from a categorical distribution over those vertices, each with probability $P_{j,k}$. After the subsequent arrival time, $T_{k+1}$, when there are $k+1$ vertices, the probability that $Z_n =j$ is 
\begin{align*}
  P_{j,k+1} = 
  \begin{cases}
	P_{j,k}(1-\Psi_{k+1}), & j \in \{1,\dots,k\} \\
	\Psi_{k+1}, & j = k+1
   \end{cases} \;.
\end{align*}
That is, the vector of probabilities $\bfP_k=(P_{1,k},\dots,P_{k,k})$ grows in length as each new vertex arrives, and each of the previous entries is scaled by $(1-\Psi_{\text{\tiny new}})$. 

\noindent\textbf{Neutrality.} The recursive scaling of $P_{j,k}$ is the essence of a neutral-to-the-left (NTL) sequence. A random vector $\mathbf{X}=(X_1,\dotsc,X_k)\in\bbR^k_+$, is NTL if the increments,
\begin{align} \label{eq:inc}
  R_j := \frac{X_j}{\sum_{i=1}^j X_i} \;,
\end{align}
form a sequence of mutually independent random variables; a non-decreasing stochastic process $Z$ defined on $\bbR$ is NTL if the vector of increments $\frac{Z(t_j) - Z(t_{j-1})}{Z(t_j)}$ is NTL for any finite partition $-\infty \leq t_1 < \dotsc < t_k \leq \infty$ of $\bbR$ \citep{Doksum:1974}. A bit of algebra shows that $\bfP_k$ is NTL: The corresponding sequence of increments is ${R_j=\Psi_j}$, for all $k$. Intuitively, this must be the case due to the recursive scaling construction. Together with the beta random variables in the sampling representation, left-neutrality characterizes these models; hence the name.

Neutral-to-the-right (NTR) processes are better known than NTL processes, and appear throughout the Bayesian statistics literature, both explicitly \citep{Walker:Muliere:1997,James:2006} and implicitly in the form of the stick-breaking constructions of the Dirichlet Process and the $\PYP$ \citep[e.g.,][]{Ishwaran:James:2001}. The properties of right- and left-neutrality are, as their names suggest, symmetric opposites: A NTR vector in reverse order is NTL, and vice versa. 

The independence properties that make NTR stick-breaking constructions useful for modeling and inference purposes transfer in large part to NTL models. The $\Psi_j$'s are conditionally independent given the $T_j$'s; along with the parameters of the beta distribution, this independence induces the factorized form in \eqref{eq:cppf}.
In the exchangeable random partitions literature, a model with joint probability that factorizes over the blocks and the probability of having $K_n$ blocks is known as \emph{Gibbs-type} \citep{Gnedin:Pitman:2006}. 

\noindent\textbf{Sparsity and power law tails in BNTL models.} The asymptotic behavior of BNTL models is controlled primarily by the arrival times, $\bfT$. In order to obtain sparse graphs, $K_n$ must be $\omega(n^{1/2})$. If $\bfT$ are the arrival times from an exchangeable sequence $\bfee$, then $K_n$ is at most $\Theta(n^{\delta})$, for some $\delta\in(0,1)$, in which case $\eta=1+\delta$ \citep{Pitman:2006}; thus, sparse graphs generated this way have $\eta\in(3/2,2)$. For the $\PYP$, $\delta=\tau$. Alternatively, for $\bfT$ sampled such that the mean interarrival time, 
\begin{align} \label{eq:ia:mean}
  \bar{\Delta}_{K_n}:=\frac{1}{K_n-1} \sum_{j=2}^{K_n}\Delta_j \;, 
\end{align}
converges to some finite $\mu$, then $K_n=\Theta(n)$ and $\eta = 1 + \frac{\mu - \alpha}{\mu - 1} > 2$. Furthermore, vertex degrees grow as $d_{j,n}=\Theta(n^{\frac{\mu - 1}{\mu - \alpha}})$ \citep{Bloem-Reddy:Orbanz:2017aa}. Thus, depending on the specification of the arrival time distribution, BNTL models can achieve any $\eta\in(1,\infty)$.

\noindent\textbf{Microclustering in BNTL partitions.} The sequence $\bfee_n$ can be transformed into an arrival-ordered partition $\randPart(\bfee_n) := \{B_{1,n},\dotsc,B_{K_n,n}\}$ of $[n]:=\{1,\dotsc,n\}$ by grouping $\bfee_n$ into blocks $B_{j,n}:=\{i\in[n] \;:\; \ee_i=j\}$. There is a bijective mapping between $\randPart(\bfee_n)$ and $G(\bfee_n)$ for all $n$ \citep{Bloem-Reddy:Orbanz:2017aa}, which puts blocks of the partition in correspondence with vertices of the graph. Hence, properties of $G(\bfee_n)$ translate into properties of $\randPart(\bfee_n)$. In particular, the growth rate of vertex degrees translates to the growth rate of blocks sizes. Recent work \citep{Betancourt:etal:2016,DiBenedetto:Caron:2017aa} has explored the so-called microclustering property, which is defined as block sizes that grow sub-linearly in $n$. The $\eta>2$ range of BNTL models corresponds precisely with this property. Although we do not make explicit statements about partition-valued data, statements about graphs are easily translated into statements about partitions via the correspondence between blocks and vertices. In particular, the inference algorithms in \cref{sec:inference} are valid for partition-valued data.

\section{INFERENCE}
\label{sec:inference}

Although PA models exhibit a range of power laws not captured by exchangeable models, they face a significant barrier to use as statistical models due to their inherent lack of exchangeability. At a high level, applying a non-exchangeable model to data for which the order is unknown requires inference over permutations of the data. This is, in general, a prohibitively difficult problem even for modest $n$. However, using the probabilistic structure of BNTL models, we design a Gibbs sampling algorithm that overcomes this difficulty for networks with thousands of vertices (\cref{sec:inference:MCMC}). If the ordered edge sequence is observed, maximum likelihood estimation scales to networks with millions of vertices (\cref{sec:mle}).

Given the hierarchical nature of BNTL models, inference may be performed at a number of levels. In the simplest case, suppose the data are a sequence of edge-ends, $\bfee_n$. From this sequence the arrival times and the arrival-ordered graph can be perfectly reconstructed, and inferring the parameters $\phi$ of the arrival distribution $\aDist^{\phi}$ and the parameters $\bfPsi_{K_n}:=(\Psi_j)_{j=1}^{K_n}$ is straightforward: Simple maximum likelihood estimators exist for $\bfPsi_{K_n}$ (see \Cref{sec:app:estimators}), and for the parameters of many arrival time distributions of interest, or equally simple MCMC samplers may be constructed for Bayesian inference.

More challenging are the situations in which some aspect of the data is not perfectly observed. For graph-valued observations, the following table summarizes the range of possibilities, in order of increasing difficulty of inference:
\begin{center}
\begin{tabular}{lll}
	\textbf{Observation} & \textbf{Unobserved variables} \\
    \midrule
    End of edge sequence $\bfee_n$ & $\alpha,\phi,\bfPsi_{K_n}$ \\
    Vertex arrival-ordered graph & $\alpha,\phi,\bfPsi_{K_n}, \bfT_{K_n}$ \\
    Unlabeled graph & $\alpha,\phi,\bfPsi_{K_n},\bfT_{K_n},\randperm [K_n]$
\end{tabular}
\end{center}
\vspace*{-0.5\baselineskip}
The last row presents a significant challenge. In particular, the unobserved variables include a permutation $\randperm$ mapping the arrival-ordered sequence to some arbitrary ordering of the vertices (by which the vertices are uniquely identified). For a graph with $K_n$ vertices, there are $K_n!$ possible permutations. Conditioned on a sequence of arrival times, some permutations have zero posterior probability, making the problem space both high-dimensional and constrained. Despite these difficulties, the inference problem is much simpler than that of a generic non-exchangeable model for a sequence of $n$ data points: Even in sparse graphs, typically $K_n \ll n$ and thus the dimension of the problem is exponentially smaller. Furthermore, the form of \eqref{eq:cppf} yields simple conditional distributions for Gibbs sampling.

\subsection{GIBBS SAMPLING}
\label{sec:inference:MCMC}

In this section, we build from the simplest inference problem to the hardest, progressing through the table in the previous section. The full sampler infers the posterior distributions of the parameters $\bfPsi_{K_n}$ and $\alpha$, of the arrival times $\bfT_{K_n}$, of the parameters of the arrival time distribution, and of the permutation of the vertices. In order to maintain the structure of the factorization over vertices in \eqref{eq:cppf}, we assume that the arrival time distribution has a Markov factorization (with a slight abuse of notation):
\begin{align}
  \boldsymbol{\aDist}^{\phi}(\bfT_k) = \delta_{T_1}(1) \prod_{j=2}^k \aDist_j^{\phi}(\Delta_j | T_{j-1}) \;,
\end{align}
with $\phi$ representing any parameters. Examples are \iid interarrivals such that $\aDist_j^{\phi}(\Delta_j|T_{j-1})=p_{\phi}(\Delta_j)$; and interarrivals that depend on the previous interarrivals through their sum and the number of previous arrivals, such as the interarrival sequence generated by exchangeable Gibbs-type sequences \citep{deBlasi:etal:2015}. 

Suppose we observe a sequence of edge-ends $\bfee_{n}$. Denote the partial sums of the ordered degree sequence as $\bar{d}_{j,n} = \sum_{i=1}^j d_{j,n}$.  
For any fixed $\alpha$ and $\phi$,
\begin{align} \label{eq:joint:sequence}
  & p_{\alpha,\phi}(\bfee_{n}, \bfPsi_{K_n}) = \\
  	& \prod_{j=2}^{K_n} \frac{\Psi_j^{d_{j,n} - \alpha - 1} (1 - \Psi_j)^{\bar{d}_{j-1,n} - (j-1)\alpha - 1}}{B(1-\alpha,\aTime_j - 1 - (j-1)\alpha)} \aDist_{\phi}(\aTime_j \mid \aTime_{j-1}) \nonumber \\
    & \quad  \times \aDist_{\phi} (T_{K_n + 1} > n \mid T_{K_n}) \nonumber \;,
\end{align}
where $B(a,b)$ is the beta function, and ${\aDist_{\phi} (T_{K_n + 1} > n \mid T_{K_n})}$ is the censored probability of vertex $K_n + 1$'s unobserved arrival time. 
Note that marginalizing $\bfPsi_{K_n}$ recovers \eqref{eq:cppf}. 

\noindent\textbf{Updates for $\bfPsi_{K_n}$.} From \eqref{eq:joint:sequence} it is clear that
\begin{align} \label{eq:psi_dist}
  \Psi_j \mid \bfee_n, \bfPsi_{\setminus j} \sim \BetaDist(d_{j,n} - \alpha, \bar{d}_{j-1,n} - (j-1)\alpha) \;,
\end{align}
where $\bfPsi_{\setminus j}$ is shorthand for the sequence $\bfPsi_{K_n}$ with $\Psi_j$ excluded. 
That is, given the arrival-ordered block sizes, the $\bfPsi_{K_n}$ are independent of each other and of the arrival times, and the beta distribution is the conjugate prior for the BNTL sampling process. To understand this, consider a second scenario in which a graph is observed with vertices labeled in order of arrival (though not their time of arrival). The data consist of an ordered sequence of degrees, $\bfn_{K_n} = (d_{1},\dotsc,d_{K_n})$, which corresponds to more than one possible edge-end sequence $\bfee_n$. The model places equal probability on each sequence that gives rise to the same arrival-ordered degree sequence $\bfn_{K_n}$ and the same arrival times; summing over these sequences yields
\begin{align} \label{eq:joint:partition}
  & p_{\alpha,\phi}(\bfn_{K_n}, \bfPsi_{K_n} \mid \bfT_{K_n},K_n) \\
  	& = \prod_{j=2}^{K_n} \frac{\Psi_j^{d_j - \alpha - 1} (1 - \Psi_j)^{\bar{d}_{j-1} - (j-1)\alpha - 1}}{B(1-\alpha,\aTime_j - 1 - (j-1)\alpha)} \binom{\bar{d}_j - T_{j}}{d_j - 1} \nonumber \;.
\end{align}
The binomial coefficients count the number of sequences $\bfee_n$ that yield $\bfn_{K_n}$, given $\bfT_{K_n}$ \citep{Griffiths:Spano:2007}. \eqref{eq:joint:partition} is a product of binomial likelihoods with beta priors. Hence, the conjugacy derived in \eqref{eq:psi_dist}. 

\noindent\textbf{Updates for $\alpha$.} In both observation scenarios, generic MCMC methods such as slice sampling \citep{Neal:2003} can be used to sample from the full conditional distribution of $\alpha$. We use slice sampling in the experiments in \cref{sec:experiments}. 

\noindent\textbf{Updates for $\phi$.} Many models of \iid interarrival times will yield conjugate updates for $\phi$. For other models, generic MCMC methods can be used. In the experiments in \cref{sec:experiments}, we consider three interarrival models: \iid $\Geom(\beta)$ and \iid $\Poisson_+(\lambda)$, which is the Poisson distribution shifted to the positive integers; and the interarrival distribution induced by the $\PYP$, which is \citep{Griffiths:Spano:2007}
\begin{align} \label{eq:arrivals:pyp}
  \aDist_{j+1}^{\theta,\tau}( & \Delta_{j+1} = s \mid T_j)  \\
    &  = (\theta + j\tau) \frac{\Gamma(\theta + T_j) \Gamma(T_j + s - 1 - j\tau)}{\Gamma(\theta + T_j + s) \Gamma(T_j - j\tau)} \nonumber \;.
\end{align}
In the former two cases, conjugate updates are performed (conditioned on $\bfT_{K_n}$); in the latter case, we perform univariate slice sampling for each of $\theta$ and $\tau$.

\noindent\textbf{Updates for $\bfT_{K_n}$.} The assumed Markov structure of the arrival times induces a simple conditional distribution for $T_j$ that is supported on the set $S_j = \{T_{j-1} + 1,\dotsc, T_{j-1} + M_j\}$, where $M_j = \min\{ T_{j+1} - T_{j-1} - 1, \bar{d}_{j-1} - T_{j-1} + 1 \}$. The support set enforces the constraints that $\bar{d}_{j-1} \geq T_j - 1$, and that $T_{j-1} < T_j < T_{j+1}$. Conditioning on $T_{j-1}$ and $T_{j+1}$, updating $T_j$ is equivalent to updating $\Delta_j$ and $\Delta_{j+1}$; for $j=2,\dots,K_n-1$, 
\begin{align*}
  & p_{\alpha,\phi} (\Delta_j = s, \Delta_{j+1}=T_{j+1}-T_{j-1}-s \mid \bfT_{\setminus j}, \bfn_n) \\
    & \propto \frac{ \aDist_{j+1}^{\phi}(T_{j+1} - T_{j-1} - s \mid T_{j-1} + s ) \aDist_j^{\phi}(s \mid T_{j-1}) }{B(1-\alpha, T_{j-1} + s - 1 - (j-1)\alpha)} \\
    & \quad \times \binom{\bar{d}_j - T_{j-1} - s}{d_j - 1} \;.
\end{align*}
For $j=K_n$,
\begin{align*}
  & p_{\alpha,\phi} (\Delta_{K_n} = s  \mid \bfT_{\setminus K_n}, \bfn) \\
    & \propto  \aDist_{K_n}^{\phi}(s \mid T_{K_n-1})\binom{n - T_{K_n-1} - s}{d_{K_n} - 1} \\
    & \quad \times \frac{ \aDist_{K_n+1}^{\phi}(\Delta_{K_n + 1} > n  - T_{K_n-1} - s \mid T_{K_n-1} - s ) }{B(1-\alpha, T_{K_n-1} + s - 1 - (K_n-1)\alpha)} \;,
\end{align*}
and $M_{K_n} = \min\{n - T_{K_n - 1} - 1, \bar{d}_{j-1} - T_{j-1} + 1 \}$.

For \iid interarrivals with distribution $p_{\phi}$, the updates are particularly easy to compute because
\begin{align} \label{eq:gibbs:iid}
  \aDist_{j+1}^{\phi}&(T_{j+1}-T_{j-1}-s\mid T_{j-1} -s) \aDist_j^{\phi}(s\mid T_{j-1}) \nonumber \\
  & = p_{\phi}(T_{j+1}-T_{j-1}-s) p_{\phi}(s) \;.
\end{align}
$p_{\phi}(s)$ can be computed for each $s \in \{1,\dotsc, M_j\}$; each term multiplied by the corresponding term in $s\in\{M_j,\dotsc,1\}$ yields \eqref{eq:gibbs:iid}. In the case of $\Geom(\beta)$ interarrivals, the distribution is uniform on $s\in\{1,\dotsc,M_j\}$:
\begin{align} \label{eq:gibbs:geom}
  \aDist_{j+1}^{\phi}&(T_{j+1}-T_{j-1}-s\mid T_{j-1} -s) \aDist_j^{\phi}(s\mid T_{j-1}) \\
  & = \beta (1-\beta)^{T_{j+1}-T_{j-1}-s-1} \beta (1-\beta)^{s-1} \propto 1 \nonumber \;.
\end{align}

\noindent\textbf{Updates for $\randperm[K_n]$.} Given a sample of $\bfT_{K_n}$, the order of the vertices can be updated via a series of adjacent swap proposals. Let $\randperm_j$ be the identity of the $j$-th vertex in the current sampling iteration. A sampling update of $\randperm$ proposes swapping $\randperm_j\leftrightarrow\randperm_{j+1}$ with probability proportional to the value of \eqref{eq:joint:partition}, with $\bfPsi_{K_n}$ marginalized and with $d_j$ and $d_{j+1}$ swapped. 
Due to the factorization over vertices, all but the $j$-th and $j+1$-st terms are the same; as a result, swap proposals are inexpensive to compute (for compactness, `$-$' indicates all other variables):
\begin{align*}
  & p_{\alpha,\phi}( \randperm_j\leftrightarrow\randperm_{j+1} | - ) \propto \frac{\Gamma(\bar{d}_{j-1} + d_{j+1}  - T_j  + 1)}{\Gamma(\bar{d}_{j+1} - d_j - T_{j+1} + 2)} \\
  & p_{\alpha,\phi}( \randperm_j\centernot\leftrightarrow\randperm_{j+1} | - ) \propto \frac{\Gamma(\bar{d}_{j-1} + d_{j}  - T_j  + 1)}{\Gamma(\bar{d}_{j} - T_{j+1} + 2)} \;.
\end{align*}
The simplicity of swap proposals enables many swaps to be sampled in a short amount of computational time, helping to overcome the high dimensionality of the sample space. We note that in general, local proposals of all possible permutations of $m>1$ consecutive vertices are possible and $m>2$ would likely enhance exploration of the state space; here we consider only $m=2$.


\begin{table*}[bt]
  \caption{Results of Gibbs sampling experiments on synthetic data $(\alpha^* = 0.75)$. The top four rows show results from each of four different BNTL models fit to a synthetic graph with 500 edges generated by the coupled $\PYP$ BNTL model; the bottom four rows show the same BNTL models fit to a synthetic graph with $\Geom(0.25)$-distributed interarrivals.}
  \label{tab:ess}
  \vspace*{-0.5\baselineskip}
  \begin{center}
  \resizebox{1.01\textwidth}{!}{
    \begin{tabular}{llllllll}
    	Gen. arrival distn. & $K_n$ & Inference model & $|\hat{\alpha} - \alpha^*|$ & $|\mathbf{\hat{S}} - \mathbf{S^*}|$ & Pred. log-lik. & Runtime (sec.) & ESS \\
    	\hline
		$\PYP(1.0,0.75)$ & $260$ & $(\tau,\PYP(\theta,\tau))$ & $\mathbf{0.046 \pm 0.002}$ & $\mathbf{28.5 \pm 0.7}$ & -$\mathbf{2637.0 \pm 0.1}$ & $297.6 \pm 0.2$ & $0.80 \pm 0.09$ \\ 
 
		$\PYP(1.0,0.75)$ & $260$ & $(\alpha,\PYP(\theta,\tau))$ & $\mathbf{0.045 \pm 0.003}$ & $33.4 \pm 1.0$ & -$2638.4 \pm 0.2$ & $313.6 \pm 0.4$ & $0.77 \pm 0.07$ \\ 
		 
		$\PYP(1.0,0.75)$ & $260$ & $(\alpha,\Geom(\beta))$ & ${0.049 \pm 0.004}$ & $66.8 \pm 1.2$ & -$2660.5 \pm 0.7$ & $90.5 \pm 0.1$ & $0.78 \pm 0.09$ \\ 
		 
		$\PYP(1.0,0.75)$ & $260$ & $(\alpha,\Poisson_+(\lambda))$ & $0.054 \pm 0.004$ & $68.0 \pm 0.7$ & -$2902.5 \pm 1.4$ & $112.5 \pm 0.1$ & $0.79 \pm 0.07$ \vspace{5pt} \\ 
		 
		$\Geom(0.25)$ & $251$ & $(\tau,\PYP(\theta,\tau))$ & $0.086 \pm 0.002$ & $56.6 \pm 1.3$ & -$2386.8 \pm 0.1$ & $295.4 \pm 0.6$ & $0.83 \pm 0.06$ \\ 
		 
		$\Geom(0.25)$ & $251$ & $(\alpha,\PYP(\theta,\tau))$ & $0.078 \pm 0.003$ & $54.2 \pm 2.0$ & -$2387.5 \pm 0.5$ & $312.7 \pm 0.3$ & $0.66 \pm 0.09$ \\ 
		 
		$\Geom(0.25)$ & $251$ & $(\alpha,\Geom(\beta))$ & $\mathbf{0.043 \pm 0.003}$ & $24.8 \pm 0.8$ & -$\mathbf{2382.6 \pm 0.2}$ & $87.2 \pm 0.1$ & $0.92 \pm 0.04$ \\ 
		 
		$\Geom(0.25)$ & $251$ & $(\alpha,\Poisson_+(\lambda))$ & $\mathbf{0.041 \pm 0.003}$ & $\mathbf{21.0 \pm 0.5}$ & -$2562.2 \pm 0.2$ & $109.5 \pm 0.1$ & $0.91 \pm 0.05$ \\ 

    \end{tabular}
  }
  \end{center}
  \vspace*{-\baselineskip}
\end{table*}

\noindent\textbf{Computational complexity.} The slice sampling updates for $\alpha$, which require evaluation of \eqref{eq:cppf}, are of complexity $\calO(K_n)$, as are the permutation swap proposals. Updates for the arrival parameter(s) $\phi$ depend on the model, but as they depend only on the $K_n$ arrival times, they are at most $\calO(K_n)$. 
The most expensive update is that of $\bfT_{K_n}$, which is $\calO(n)$, though the constant hidden in $\calO$ may vary greatly across arrival models.

\subsection{MAXIMUM LIKELIHOOD FOR PARAMETERS IN EDGE SEQUENCES}
\label{sec:mle}
Suppose the edge-end sequence $\bfee_{n}$ is observed. For arrival time distribution $\aDist^{\phi}$, $\phi$ and $\alpha$ can be estimated by maximum likelihood (ML). The likelihood admits the factorization
\begin{align} \label{eq:likelihoodfactorization}
p_{\alpha,\phi}(\bfee_n) &= p_{\alpha}(\bfee_n|\bfT_{K_n})\aDist_{\phi}(\bfT_{K_n}) \;,
\end{align}
with the practical implication that the estimating equations for $\alpha$ and $\phi$ can be solved separately. In particular, 
\begin{align} \label{eq:alpha:mle}
  \hat{\alpha} = \argmax_{\alpha\in (-\infty,1)} \log p_{\alpha}(\bfee_n|\bfT_{K_n}) \;,
\end{align}
where $p_{\alpha}(\bfee_n \mid \bfT_{K_n})$ is as in \eqref{eq:cppf}. 

Closed-form MLEs are known for many \iid interarrival distributions. For the $\Geom(\beta)$ and $\Poisson_+(\lambda)$ distributions used in \cref{sec:experiments}, $\hat{\beta} = \frac{K_n-1}{n - K_n}$ and $\hat{\lambda} = \frac{n - K_n}{K_n - 1}$. MLEs for $\theta$ and $\tau$ in $\PYP$-induced interarrivals can be found by numerically optimizing the product over arrival times of \eqref{eq:arrivals:pyp}. See \Cref{sec:app:mle} for details. Maximum a posteriori (MAP) estimates are straightforward to compute by placing priors on the model parameters and including the relevant prior probabilities in \eqref{eq:likelihoodfactorization}-\eqref{eq:alpha:mle}.

\subsection{RELATED WORK}
\label{sec:inference:related}

There is relatively little previous work on statistical inference for non-exchangeable models of network data. \Citet{Bloem-Reddy:Orbanz:2018} develop sequential Monte Carlo methods for non-exchangeable models; those methods are feasible only for networks with hundreds of vertices. See references therein for related ideas based on importance sampling. Where BNTL models overlap with edge exchangeable models, there exist inference algorithms that do not account for arrival times. Namely, if $\bfee_n$ is assumed to be an exchangeable sequence of edge-ends, then the sampling and estimation algorithms for Gibbs-type partitions can be used. For example, Gibbs sampling methods for the $\PYP$ are derived in \citet{Ishwaran:James:2001}. \Citet{Crane:Dempsey:2017} give maximum likelihood estimating equations. However, neither method infers arrival times, and the inference techniques do not extend to the wider class of non-exchangeable BNTL models.

\Citet{Wan:2017} propose MLEs for the parameters of a class of PA models when the edge sequence is observed. A MLE of the parameter $\alpha$ in a slightly different PA model was proposed by \citet{Gao:2017} for observed edge sequences. The PA model considered there has random initial degrees, rather than random arrival times, but the initial degrees play a similar role to the arrival times. Those authors find that conditioned on the initial degrees, the degree sequence at step $n$, $\mathbf{d}_{K_n}$, is sufficient for $\alpha$, and that the MLE is asymptotically normal. Based on the similarities of the models and the corresponding log-likelihoods, it is plausible that similar properties hold for BNTL models.

\section{EXPERIMENTS}
\label{sec:experiments}

We apply the inference methods developed in \cref{sec:inference} to data. The first set of experiments is in the unlabeled network setting, in which the posterior distribution over vertex ordering must be inferred along with the model parameters. 
In a second set of experiments, we consider graphs with edges labeled in order of appearance, and demonstrate that maximum likelihood and MAP estimation scale to networks with millions of nodes.\footnote{Julia code is available at \url{https://github.com/emilemathieu/NTL.jl}.}

\subsection{BAYESIAN INFERENCE}
\label{sec:inf:bayes}

We first apply the Gibbs sampler from \cref{sec:inference:MCMC} to synthetic data, which allows us to study the effects of model misspecification on parameter estimation, and to demonstrate the feasibility of inference over the vertex order and the arrival times. We generated two synthetic graphs, each with 1,000 edges: One from a $\PYP(\theta,\tau)$ sequence \eqref{eq:pyp} in which $\tau$ is forced to be equal to the BNTL parameter $\alpha$, which corresponds to the edge exchangeable Hollywood model of \citet{Crane:Dempsey:2017}; and one from a BNTL model with \iid $\Geom(\beta)$-distributed interarrival times. We set $\theta=1.0$, $\beta=0.25$, and in both cases, $\alpha=0.75$. 

For each of the graphs, we held out the final 500 edges for prediction, and we fit four different BNTL models to the first 500 edges whose order we treated as unknown: One with $\PYP(\theta,\tau)$-induced arrivals and $\alpha=\tau$ (the ``coupled $\PYP$'' model), which is the same as the generative model of the first synthetic dataset; one with $\PYP(\theta,\tau)$-induced arrivals and $\alpha$ allowed to vary separately from $\tau$ (the ``uncoupled $\PYP$'' model); and two \iid interarrival models, with $\Geom(\beta)$- and $\Poisson_+(\lambda)$-distributed interarrivals. We ran 125,000 Gibbs sampling iterations, including a burn-in of 25,000, and collected one in every 1,000 iterations for a total of 1,000 samples. To assess performance, we calculated the average absolute error (relative to the true value) of MCMC samples of $\alpha$, and of $\mathbf{S}:= \frac{1}{K_n-1}\sum_{j>1} (\bar{d}_{j-1} - T_j)$. The latter statistic captures how well the sampler recovers the vertex permutation and the arrival times. We also calculated the predictive log-likelihood of a further 500 edges. Average runtimes\footnote{All Gibbs sampling experiments were run on a quad-core (3.1 GHz) Dell desktop running Linux.} and effective sample size (ESS) factors, based on the log of the normalized $L_1$ distance between the sampled degree sequence and the true degree sequence, are also shown.

\begin{table}[t]
  \caption{Scaling performance of the Gibbs sampler.}
  \label{tab:ess:scale:n}
  \vspace*{-0.5\baselineskip}
  \begin{center}
  \resizebox{0.485\textwidth}{!}{
  	\begin{tabular}{l  lll}
  		 & 100 edges & 1,000 edges & 10,000 edges \\ 
 		\hline
		$|\hat{\alpha} - \alpha^*|$ & $0.12 \pm 0.01$ & $0.03 \pm 0.00$ & $0.01 \pm 0.00$ \\ 
		 
		$|\hat{\beta} - \beta^*|$ & $0.02 \pm 0.00$ & $0.01 \pm 0.00$ & $0.00 \pm 0.00$ \\ 
		 
		$|\hat{\mathbf{S}} - \mathbf{S}^*|$ & $10.3 \pm 0.4$ & $33.9 \pm 0.9$ & $343 \pm 1.6$ \\ 
		
		ESS		& $0.90 \pm 0.04$ & $0.85 \pm 0.05$ & $0.75 \pm 0.08$ \\ 

		Runtime (s) & $21 \pm 0.0$ & $213 \pm 0.4$ & $2267 \pm 2$ \\ 

  	\end{tabular}
  }
  \end{center}
  \vspace*{-\baselineskip}
\end{table}

\Cref{tab:ess} summarizes the results, averaged over 10 repetitions. The top four rows show the results of fitting four BNTL inference models to the coupled $\PYP$ dataset. Unsurprisingly, the inference models with arrivals induced by the $\PYP$ achieve the lowest errors in $\alpha$ and $\mathbf{S}$, and highest predictive log-likelihood. The bottom four rows show the same four inference models fit to the $\Geom(0.25)$ BNTL dataset; the \iid interarrival models achieve lower errors, and the $\Geom(\beta)$ inference model attains the highest predictive log-likelihood. Although the $\Poisson_+(\lambda)$ inference model attains low errors in $\alpha$ and $\mathbf{S}$, the low variance of the Poisson distribution compared to the Geometric distribution means that it attains low predictive probability due to the relatively frequent occurrence of large interarrivals.

As discussed in \cref{sec:inference:MCMC}, the most expensive Gibbs update is that of the arrival time sequence. As such, the $\Geom(\beta)$ interarrival inference model benefits greatly from \eqref{eq:gibbs:geom}, which implies that computation of $\aDist_j^{\phi}$ is not required. The \iid interarrival models each have conjugate updates for their parameters, whereas the $\PYP$ interarrival models require slice sampling for $\phi=(\theta,\tau)$. These differences are reflected in the runtimes shown in \cref{tab:ess}. Finally, all four inference models exhibit good ESS factors, indicating that the sampler is exploring permutation space beyond simply swapping vertices of the same degree.

\begin{table}[b]
\vspace*{-\baselineskip}
\caption{SNAP temporal network datasets.}
\label{tab:datasets}
\begin{center}
	\begin{tabular}{lll}
		Dataset                 & \# of vertices   & \# of edges    \\
		\hline
		Ask Ubuntu    & 159,316   & 964,437    \\
		UCI social network   & 1,899     & 20,296     \\
		EU email        & 986       & 332,334    \\
		Math Overflow & 24,818    & 506,550    \\
		Stack Overflow           & 2,601,977 & 63,497,050 \\
		Super User    & 194,085   & 1,443,339  \\
		Wikipedia talk pages    & 1,140,149 & 7,833,140 \\
	\end{tabular}
\end{center}
\end{table}

\noindent\textbf{Scaling in $n$.} In order to study how sampling and computational efficiency scale with the size of the network, we generated a single BNTL network of 10,000 edges with \iid $\Geom(0.25)$-distributed arrival times, and performed Gibbs sampling using the subgraphs formed by the first $n$ edges, with $n\in\{100, 1,\!000, 10,\!000\}$. \Cref{tab:ess:scale:n} shows the results of 10 repetitions, each of 150,000 Gibbs iterations; samples were collected once every 1,000 iterations after a burn-in period of 75,000 iterations. Parameter estimation is increasingly accurate for increasing $n$ without major decrease in ESS, indicating that the sampler is taking advantage of the increased statistical signal in the bigger network. Runtimes increase at a rate linear in $n$.

\begin{table*}[!ht]
\caption{MLEs on full datasets, and predictive log-likelihood for final 20\% of edges based on MLEs fit to the first 80\%, for three different BNTL models. Note that the uncoupled $\PYP(\theta,\tau)$ and $\Geom(\geom)$ interarrival models have the same $\hat{\alpha}$ due to the factorization in \eqref{eq:likelihoodfactorization}.}
\label{tab:mle_diagnostics}
\vspace*{-0.5\baselineskip}
\begin{center}
\resizebox{\textwidth}{!}{
\begin{tabular}{llll | lllllll}
\multirow{2}{*}{Dataset} & \multicolumn{3}{c}{Coupled $\PYP(\theta,\alpha)$}  & &  \multicolumn{2}{c}{Uncoupled $\PYP(\theta,\tau)$} &  & \multicolumn{3}{c}{$\Geom(\geom)$}                             \\
\cline{2-4} \cline{6-7} \cline{9-11}
                         & $(\hat{\theta},\hat{\alpha})$ & $\hat{\eta}$           & Pred. l-l.            & $\hat{\alpha}$    & $(\hat{\theta},\hat{\tau})$ & Pred. l-l.     &     & $\hat{\beta}$  & $\hat{\eta}$   & Pred. l-l.                             \\
\hline
Ask Ubuntu  & (18080, 0.25) & 1.25   & -3.707e6              & -2.54            & (-0.99, 0.99)    & -3.678e6  &   &  0.083 & 2.32    & \textbf{-3.678e6}                         \\
UCI social network & (320.4, 4.4e-11) & --   & -1.600e5            & -4.98             & (5.50, 0.52)     & \textbf{-1.595e6}   &  & 0.016  & 2.10  & -1.596e5                    \\
EU email   & (113.6, 2.5e-14)  & --    & \textbf{-8.06e5}              & -1.86              & (113.6, 9.2e-10)      & \textbf{-8.06e5}  &   & 0.001  & 2.00   & -8.07e5                     \\
Math Overflow & (2575, 0.15) & 1.15  & -1.685e6              & -6.62           & (-0.97, 0.997)    & -1.670e6 &    & 0.025  & 2.19   & \textbf{-1.670e6}                \\
Stack Overflow & (297600, 0.11) & 1.11 & -3.358e8             & -8.94            & (-1.0, 1.0)      &  -3.333e8  &    &  0.020  & 2.21   & \textbf{-3.333e8}                 \\
Super User  & (20640, 0.24) & 1.24  & -5.855e6            & -4.19          & (-0.996, 1.0)      &  \textbf{-5.775e6}   &  & 0.067 & 2.37    & -5.775e6              \\
Wikipedia talk pages  & (14870, 0.54) & 1.54  & -3.074e7          & -0.25            & (-1.0, 1.0)   & \textbf{-3.066e7}  &   & 0.073  & 2.10   & -3.066e7                   
\end{tabular}
}
\end{center}
\vspace*{-\baselineskip}
\end{table*}

\subsection{MAXIMUM LIKELIHOOD ESTIMATION ON EDGE SEQUENCES}
\label{sec:experiments:mle}

For observed edge sequences, maximum likelihood estimation scales to networks with millions of vertices and tens of millions of edges. To demonstrate, we compute MLEs on a collection of temporal network datasets available from the Stanford Network Analysis Project (SNAP) \citep{snapnets}. 

\begin{figure}[!hb]
	\begin{center}
		\includegraphics[width=0.23\textwidth]{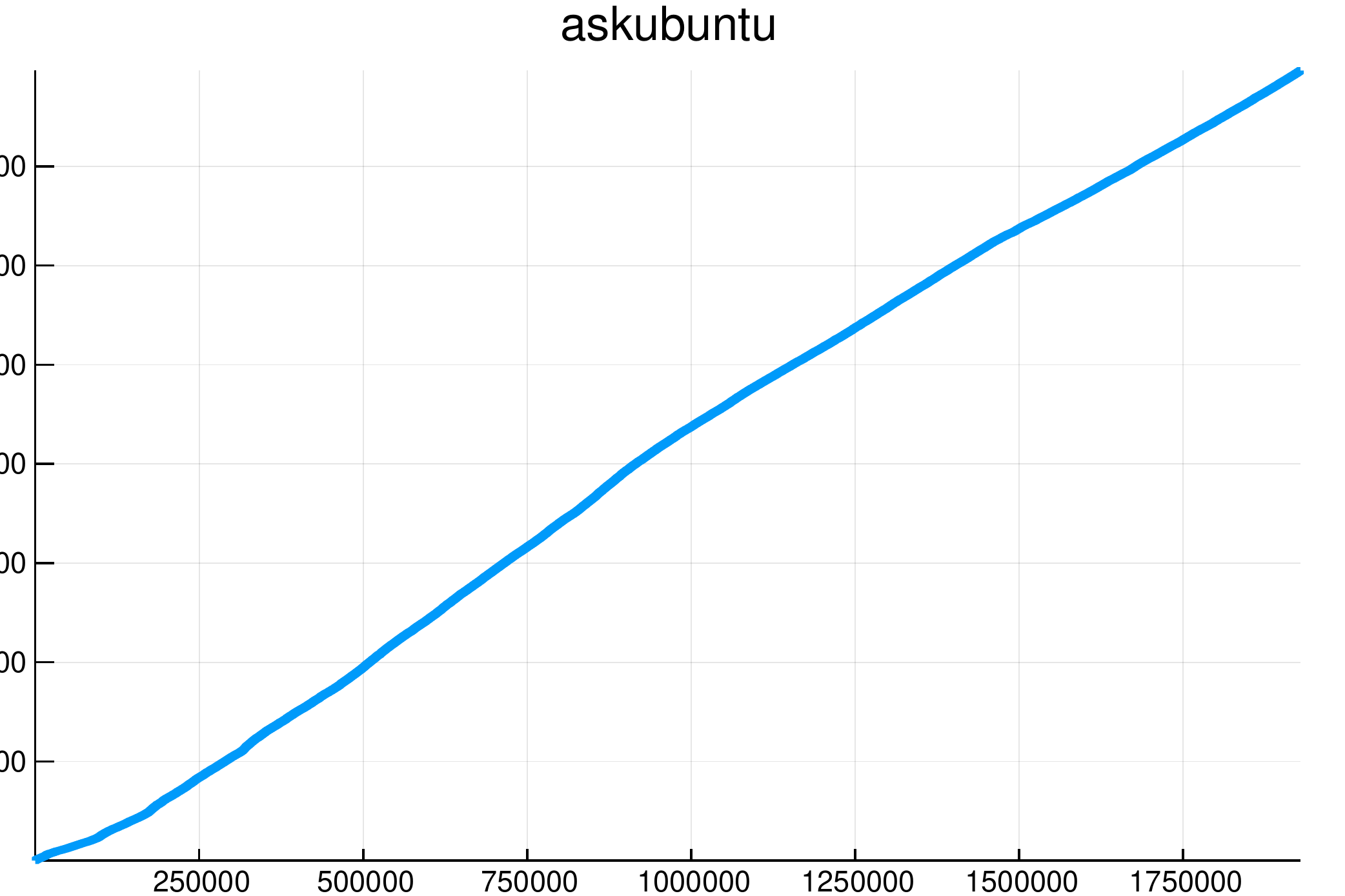}
		\includegraphics[width=0.23\textwidth]{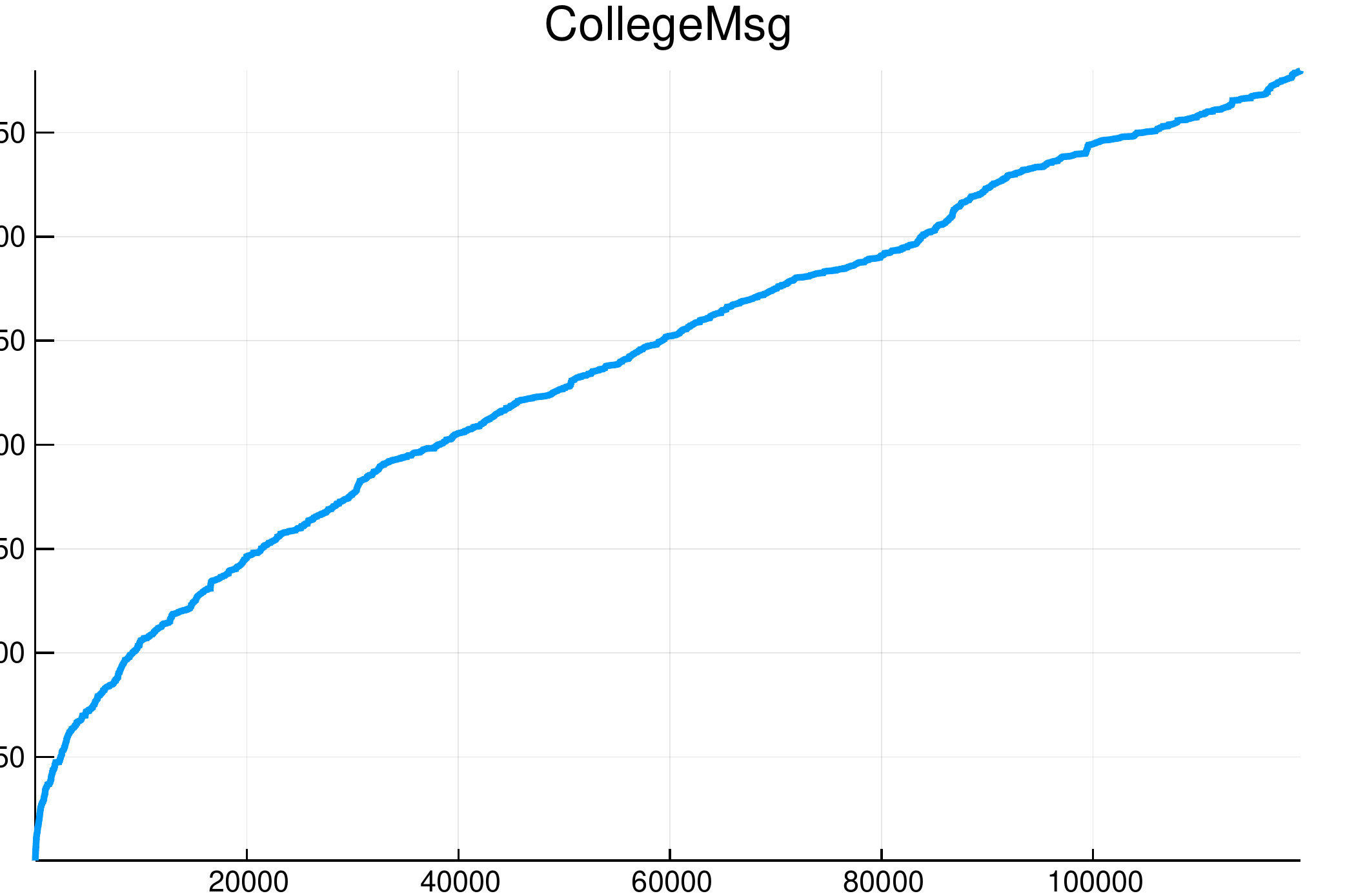}
		\vspace{8pt} \\
		\includegraphics[width=0.23\textwidth]{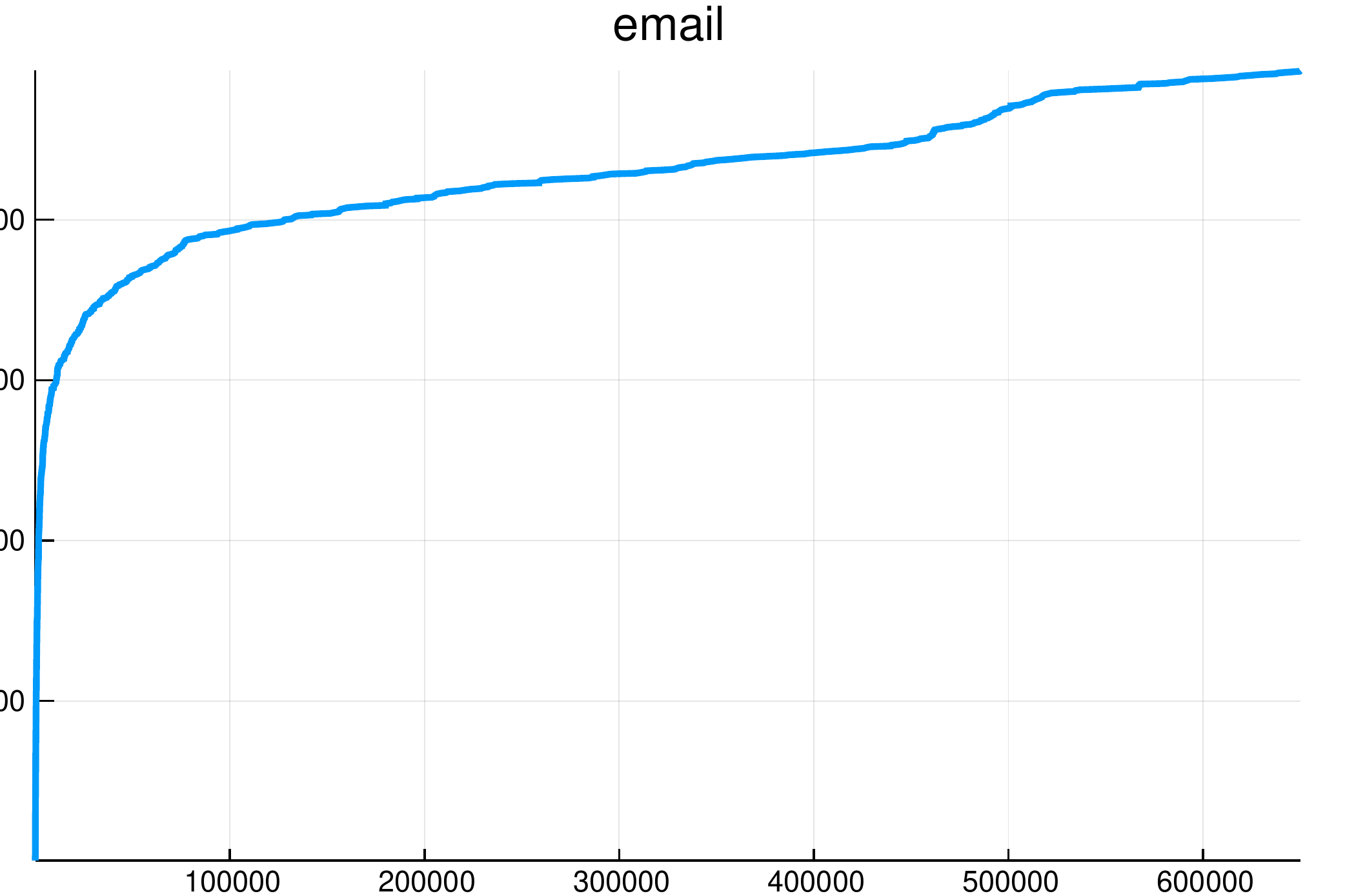}
		\includegraphics[width=0.23\textwidth]{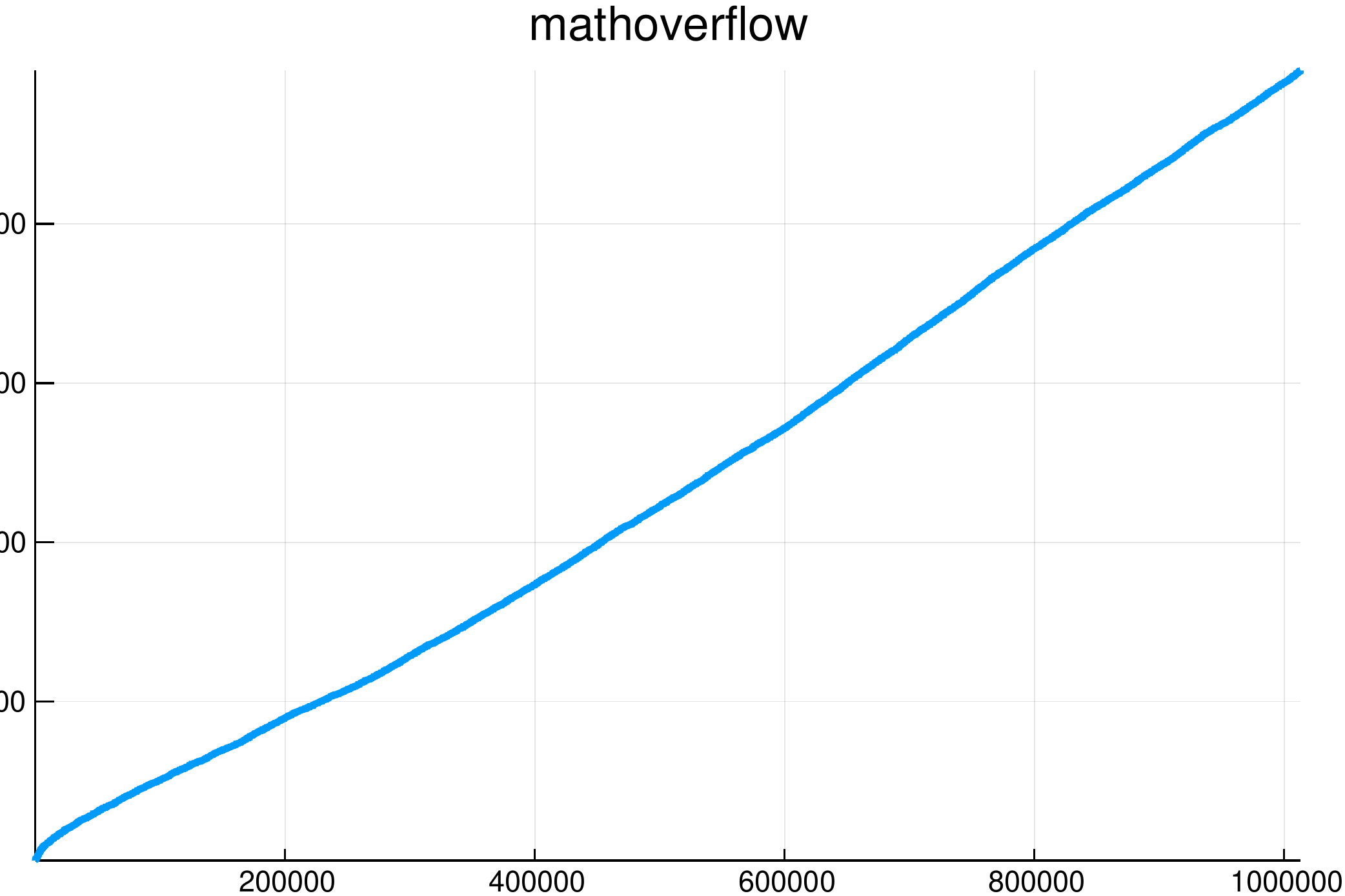}
		\vspace{8pt} \\
		\includegraphics[width=0.23\textwidth]{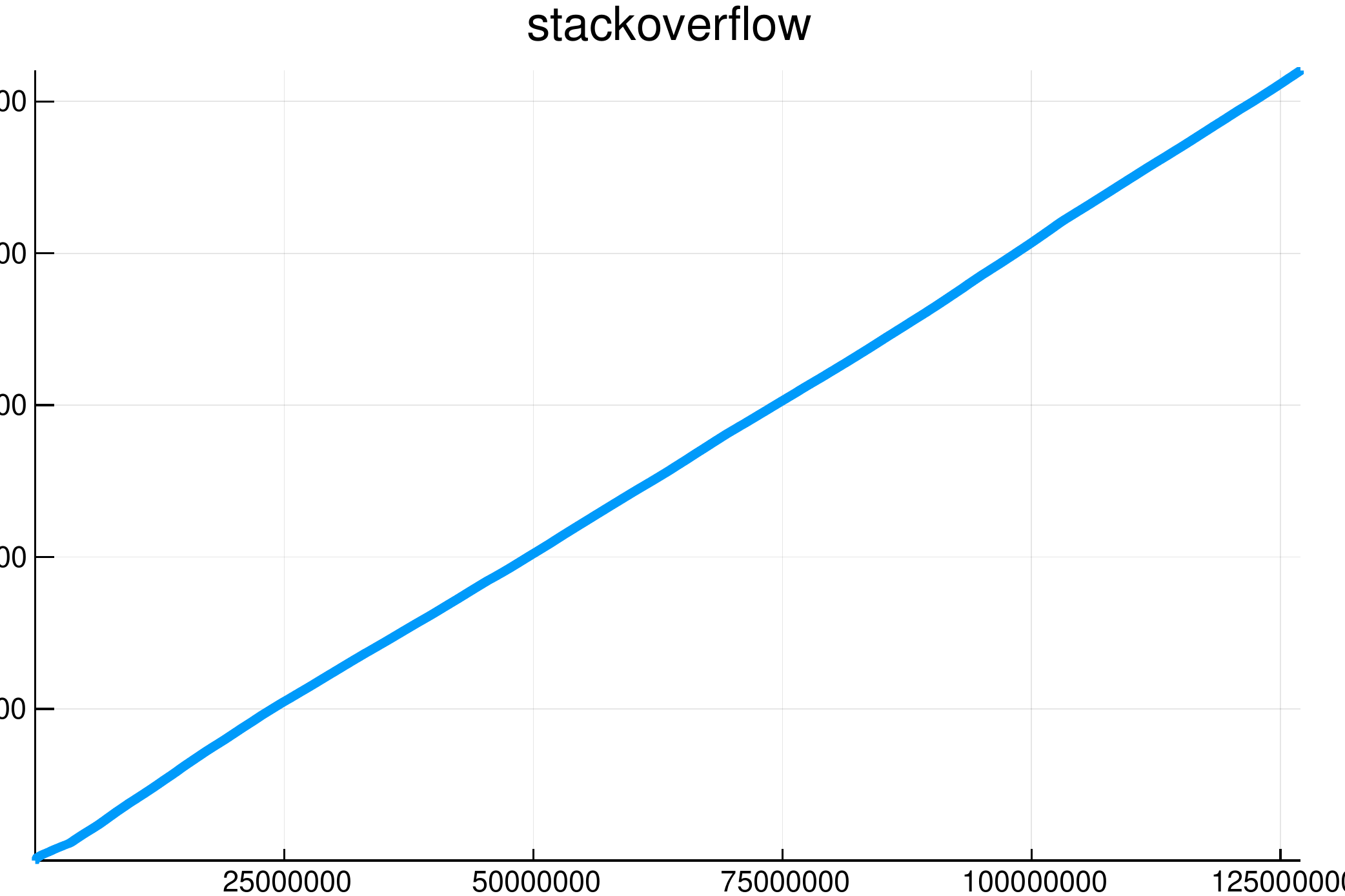}
		\includegraphics[width=0.23\textwidth]{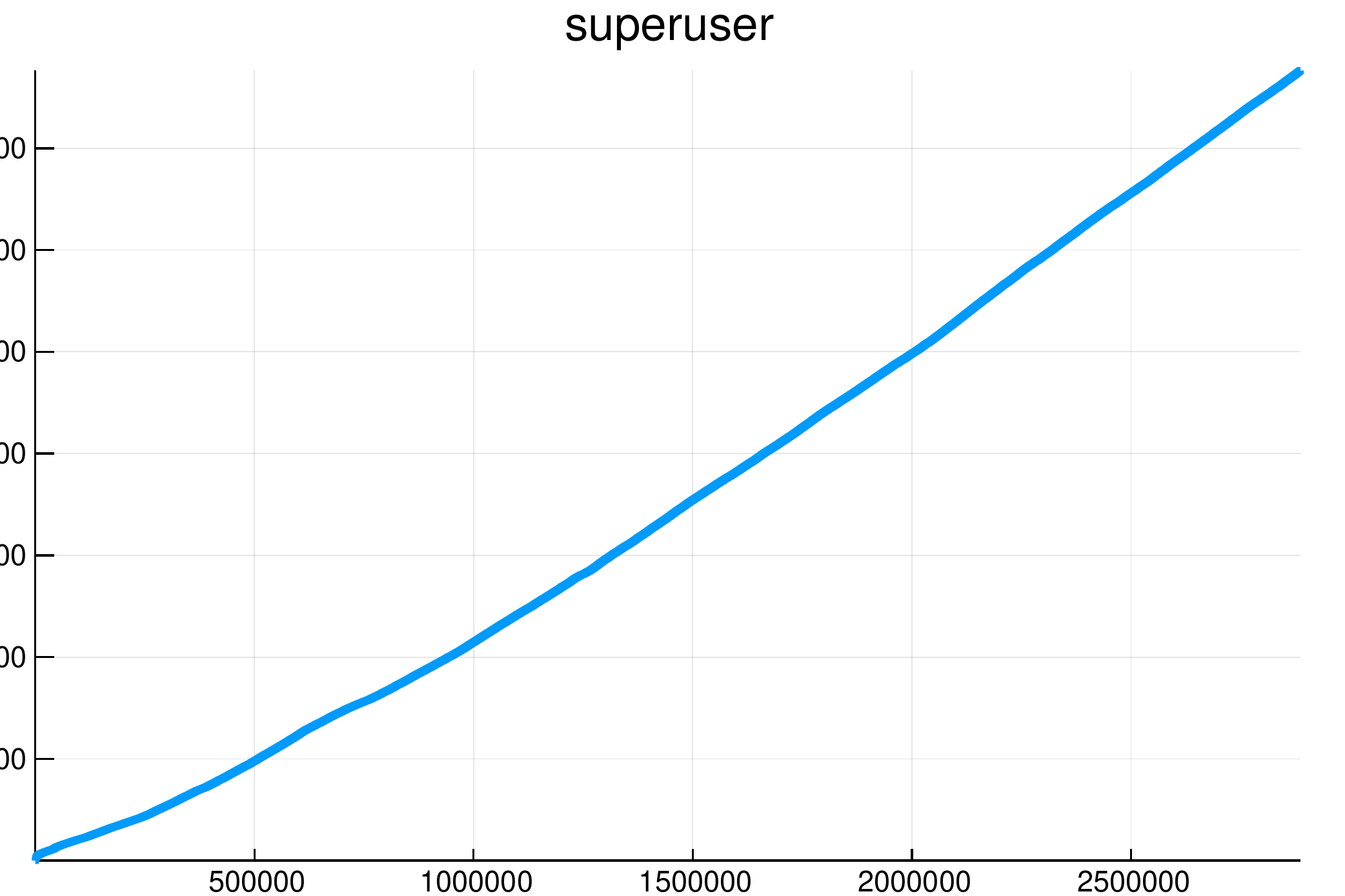}
		\vspace{8pt} \\
		\includegraphics[width=0.23\textwidth]{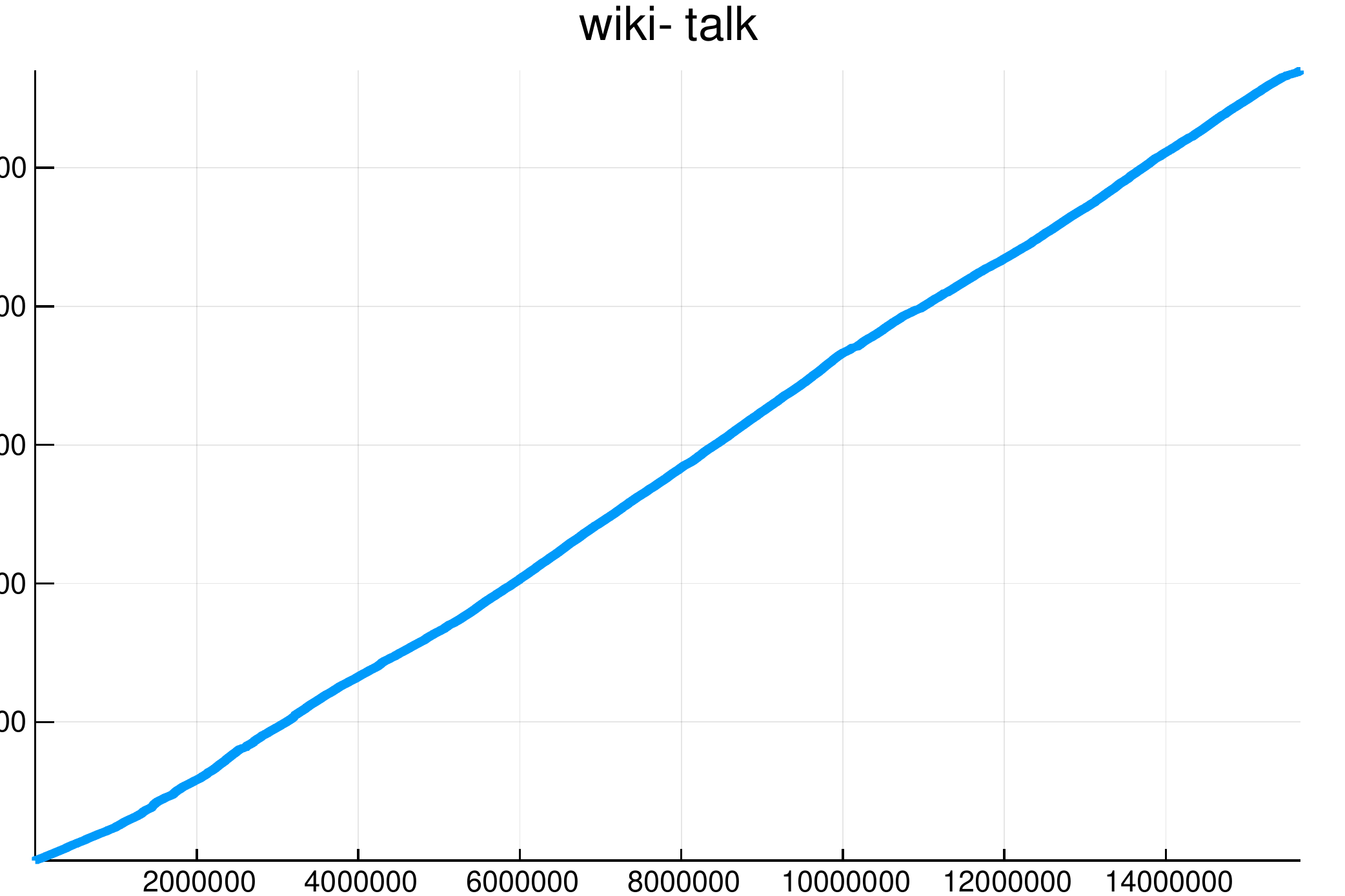}
	\end{center}
	\caption{Arrival time sequences for SNAP data. Vertical axis is number of vertices, $K_n$; horizontal axis is $n$.}
	\label{fig:snap:arrivals}
\end{figure}

For each of the datasets listed in \cref{tab:datasets}, we fit MLEs of $\alpha$ and of the parameters of three different interarrival models: coupled $\PYP(\theta,\alpha)$; uncoupled $\PYP(\theta,\tau)$; and $\Geom(\beta)$. \Cref{tab:mle_diagnostics} displays the MLEs of the model parameters and the plug-in estimates of the asymptotic power law degree exponent, $\eta$. (The asymptotic degree distribution of the uncoupled $\PYP$ model is unknown.) Note that due to the factorization of the likelihood in \eqref{eq:likelihoodfactorization}, $\hat{\alpha}$ is the same for any model in which $\alpha$ is not coupled to the arrival distribution. In order to assess model fitness, we fit MLEs for the same BNTL models to the first 80\% of the edges in each network and calculated the predictive log-likelihood based on the MLEs of the remaining 20\%; this is also shown in \cref{tab:mle_diagnostics}. 
For context, the arrival time sequence of each dataset is plotted in \cref{fig:snap:arrivals}. Unsurprisingly, whether or not the arrival times are approximately linear in $n$ largely determines which BNTL model fits best. The two densest networks, the EU email and UCI social networks, exhibit arrival times that are sub-linear in $n$; as such, the $\PYP$ models fit best. Note that the coupled $\PYP$ model estimates $\hat{\alpha}\approx 0$, indicating the lack of a power law tail in the degree distribution. In the rest of the networks, the arrival times appear approximately linear in $n$. The $\Geom(\geom)$ and uncoupled $\PYP$ model fit best. However, we note that in these cases the MLEs for the uncoupled $\PYP$ model are at the boundaries of the parameter range $(\hat{\theta}\approx -1, \hat{\tau}\approx 1)$. This illustrates that although the uncoupled $\PYP$ model is more flexible than the coupled version, the underlying arrival time model cannot capture linear arrival time sequences without driving the parameters to the boundaries.

\section{DISCUSSION}
\label{sec:discussion}

BNTL models are a useful tool to reason about asymptotic properties of a network. For example, the exponent of the asymptotic power law tail is a function of model parameters, which can be estimated from finite-size networks without dealing with the large fluctuations of heavy-tailed degree distributions in finite samples. Furthermore, the ability to capture the full range of power law exponents and sparsity levels within the same model class allows for model fitness comparisons using the same set of techniques, as in \cref{sec:experiments}. 
We have designed a set of inference algorithms for these models; in doing so, we have made a large class of previously intractable models useful for statistical inference. 

\noindent\textbf{Future research directions.} The full Gibbs sampler scales reasonably well to networks with thousands of vertices; in order to scale to larger networks, further work is needed. One possible approach is via Metropolis--Hastings with cheap joint proposals of the arrival times and the permutation, which may be able to take larger steps in sample space. A different direction is variational inference, though permutations pose a significant challenge in that context; recent work \citep{Linderman:etal:2017} is a step in that direction.

\newpage

\subsubsection*{Acknowledgments}

\vspace{-0.25\baselineskip}
BBR, EM, YWT's research leading to these results received funding from the
European Research Council under the European Union's Seventh Framework
Programme (FP7/2007-2013) ERC grant agreement no.~617071. EM, YWT acknowledge Microsoft Research and EPSRC for
partially funding EM's studentship. AF acknowledges funding from EPSRC grant no.~EP/N509711/1.

\bibliography{references}

\clearpage

\appendix

\crefalias{section}{appsec} 

\section{DEGREE DISTIRBUTIONS WITH POWER LAW TAILS}
\label{sec:app:power:laws}

Let $D$ be a random variable from some distribution $P = (p_d)_{d\geq 1}$ with power law tails. Then
\begin{align*}
  \bbE[D] = \sum_{d\geq 1} d \cdot p_d = C \cdot \sum_{d\geq d^*} d \cdot d^{-\eta} \;,
\end{align*}
for some constants $C,d^*$ that control the tail approximation. The sum of terms $d^{-(\eta-1)}$ converges if and only if $\eta > 2$. 

For a graph $G_n$, the average degree is
\begin{align} \label{eq:avg:deg:def}
  \bar{d}_n = \frac{1}{K_n} \sum_{j=1}^{K_n} d_{j,n} = \frac{2n}{K_n} \;.
\end{align}
Let $D_n$ be the degree of a vertex sampled uniformly at random from $G_n$. If $m_d(n)/K_n \xrightarrow[]{\text{p}} p_d$ for all $d\in\bbN$, then $D_n \xrightarrow[]{\text{d}} D$ as $n\to\infty$ and $\bar{d}_n \xrightarrow[]{\text{a.s.}} \bbE[D]$. 

If $K_n = o(n)$, then by \eqref{eq:avg:deg:def} $\bar{d}_n\to \infty$, which implies that $\bbE[D]=\infty$. On the other hand, if $K_n = \Theta(n)$, then $\bar{d}_n \to \bbE[D] < \infty$.

The \emph{Fact} in \cref{sec:powerlaws} is an assertion of these property. 

\section{UNBOUNDED AVERAGE DEGREE IN EXCHANGEABLE POINT PROCESS MODELS}
\label{sec:app:graphex}

As with edge exchangeable models, models based on exchangeable point processes have unbounded expected average degree. We refer the reader to  \citet{Caron:Fox:2017,Veitch:Roy:2015,Borgs:etal:2016} for details on such models.  
Ignoring self-loops, the degree $D_{\nu}(\lambda)$ of a fixed vertex (with ``position'' $\lambda\in\bbR_+$) is $\Poisson(\nu \mu_W(\lambda))$, where $\nu$ is the size parameter of the point process \citep[][Lemma 5.1]{Veitch:Roy:2015}; taking $\nu\to\infty$ yields the asymptotic properties, and for non-trivial $\mu_W$ (i.e., those that generate sparse graphs), $\lim_{\nu\to\infty} \bbE[D_{\nu}(\lambda)]=\infty$ for all $\lambda$. 

\section{ESTIMATORS FOR $\Psi_j$}
\label{sec:app:estimators}

When the arrival times are known, it is straightforward to show that the MLE for $\Psi_j$ is
\begin{align}
  \hat{\Psi}^{\text{\tiny MLE}}_j = \frac{d_{j,n} - 1}{\bar{d}_{j,n} - T_j} \;.
\end{align}
If only the arrival order is observed, then the maximum a posteriori estimator (MAPE) corresponding to $\Psi_j$ is
\begin{align}
  \hat{\Psi}^{\text{\tiny MAP}}_j = \frac{d_j - 1 - \alpha}{\bar{d}_j - j\alpha - 2} \;.
\end{align}
Note that the MAPE does not require knowledge of the arrival times, but requires specification of $\alpha$. A consistent estimator that depends neither $\alpha$ nor the arrival times is given by \citep{Bloem-Reddy:Orbanz:2017aa}
\begin{align} \label{eq:beta:convergence}
  \frac{d_j}{\bar{d}_j} \xrightarrow[n\to\infty]{\text{\small a.s.}} \Psi_j \quad \text{for all} \quad j \geq 1 \;.
\end{align}

\section{DETAILS OF MLEs FOR $\PYP$ AND GEOMETRIC INTERARRIVALS}
\label{sec:app:mle}
Starting from equation (14) in the main text, the likelihood of observed data $\bfee_n$ with degree distribution $m_n(d) := \#\{j: d_{j,n} = d\}$ under a BNTL model with inter-arrivals Geom($\beta$) and NTL parameter $\alpha$ is
\begin{align}
	\bbP[\bfee_n|\geom, \alpha] &=&& \geom^{(K_n-1)}(1-\geom)^{(n-K_n-1)} \nonumber \\
    &&& \times \prod_{i\not\in \bfT_{K_n}}\frac{d_{z_i, i} - \alpha}{i - 1 - K_{i-1}\alpha}
\end{align}
Observe that
\begin{align}
	\prod_{i \not \in \bfT_{K_n}} (d_{z_i, i} - \alpha) &= \prod_{j=1}^{K_n}\prod_{i=2}^{d_{j,n}} (i-1-\alpha) \nonumber \\
    &= \prod_{j=1}^{K_n} \frac{\Gamma(d_{j,n} - \alpha)}{\Gamma(1-\alpha)} \nonumber \\
    &= \Gamma(1-\alpha)^{-K_n}\prod_{d=1}^\infty \Gamma(d-\alpha)^{m_n(d)}
\end{align}
This yields
\begin{align}
    \bbP[\bfee_n|\geom, \alpha] &=&& \geom^{(K_n-1)}(1-\geom)^{(n-K_n-1)} \nonumber \\
    &&& \times \Gamma(1-\alpha)^{-K_n}\prod_{d=1}^\infty \Gamma(d-\alpha)^{m_n(d)} \nonumber \\
    &&& \times \prod_{i\not\in \{T_j\}} (i - 1 - K_{i-1}\alpha)^{-1}
\end{align}

For the coupled $\PYP$, the BNTL parameter $\alpha=\tau$ is coupled to the arrival process. The factorization (14) is less helpful. The full likelihood in this case is
\begin{align}
	\bbP[\bfee_n|\theta, \tau] &=&& \prod_{j=1}^{K_n} \frac{\theta + j\tau}{T_j - 1 +\theta} \nonumber \\
    &&& \times \prod_{i\not\in \bfT_{K_n}} \frac{d_{z_i, i} - \tau}{i - 1 - \theta} \\
    &=&& \frac{\Gamma(1+\theta)}{\Gamma(n+\theta)} \; \prod_{j=1}^{K_n}   (\theta+j\tau)  \nonumber \\
    &&& \times \Gamma(1-\tau)^{-K_n}\prod_{d=1}^\infty \Gamma(d-\tau)^{m_n(d)}
    \label{eq:pypcoupledlikelihood}
\end{align}

Finally, for the uncoupled $\PYP$ in which $\alpha$ and $\tau$ are independent parameters, we again make use of (14) to write the likelihood as 
\begin{align}
    \bbP[\bfee_n|\geom, \alpha] &=&& \frac{\Gamma(1+\theta)}{\Gamma(n+\theta)} \prod_{j=1}^{K_n} (\theta + j\tau) \nonumber \\
    &&& \times \prod_{i\not \in \bfT_{K_n}} (i - 1 - K_{i-1}\tau) \nonumber \\
    &&& \times \Gamma(1-\alpha)^{-K_n}\; \prod_{d=1}^\infty \Gamma(d-\alpha)^{m_n(d)} \nonumber \\
    &&& \times \prod_{i\not\in \bfT_{K_n}} (i - 1 - K_{i-1}\alpha)^{-1}
\end{align}
and one can readily see that setting $\alpha=\tau$ reduces to \eqref{eq:pypcoupledlikelihood}.

\end{document}